\documentclass{article} 
\usepackage{iclr2020_conference,times}

\usepackage{amsmath,amsfonts,bm}

\def\eqref#1{equation~\ref{#1}}

\def\1{\bm{1}}

\DeclareMathAlphabet{\mathsfit}{\encodingdefault}{\sfdefault}{m}{sl}
\SetMathAlphabet{\mathsfit}{bold}{\encodingdefault}{\sfdefault}{bx}{n}

\usepackage{hyperref}
\usepackage{url}

\usepackage{verbatim}
\usepackage{xspace} 
\usepackage{graphicx}
\usepackage{adjustbox}
\usepackage{array}
\usepackage{multirow}
\usepackage{booktabs}
\usepackage{caption}
\usepackage{subcaption}
\usepackage{color, colortbl}
	
\definecolor{green}{rgb}{0.5,1,0.5}
\definecolor{orange}{rgb}{1,0.7,0.3}
\definecolor{red}{rgb}{1.0,0.5,0.5}

\usepackage{txfonts}

\title{Poly-encoders: architectures and pre-training \\strategies for fast and accurate multi-sentence scoring}

\author{Samuel Humeau\thanks{\,\,Joint First Authors.}\,, Kurt Shuster$^{*}$, Marie-Anne Lachaux, Jason Weston\\
Facebook AI Research \\
{\tt \{samuelhumeau,kshuster,malachaux,jase\}@fb.com}
}

\iclrfinalcopy 
\begin{document}

\maketitle

\begin{abstract}
The use of deep pre-trained 
transformers has led to remarkable progress in 
a number of applications \citep{devlin-etal-2019-bert}. 
For tasks that make pairwise comparisons between sequences,  matching a given input with a corresponding label,
two approaches are common: 
 \textit{Cross-encoders} performing full self-attention over the pair and   \textit{Bi-encoders} encoding the pair separately. The former often performs better, but is too slow for practical use.
In this work, we develop a new transformer architecture, the \textit{Poly-encoder}, that learns global rather than token level self-attention features. 
%
We perform a detailed comparison of all three approaches, including 
what pre-training and fine-tuning strategies work best. 
We show  our models achieve state-of-the-art results on four tasks;
that Poly-encoders are faster than Cross-encoders and more accurate than Bi-encoders;
and that the best results are obtained by pre-training on large datasets
similar to the downstream tasks. 

\end{abstract}

\section{Introduction}

Recently, substantial improvements to state-of-the-art benchmarks on a variety of language understanding tasks have been achieved through the use of deep pre-trained language models followed by fine-tuning \citep{devlin-etal-2019-bert}. 
In this work we explore improvements to this approach for the class of tasks 
that require multi-sentence scoring: given an input context, score a set of candidate labels, a setup common in retrieval and dialogue tasks, amongst others. Performance in such tasks has to be measured via two axes: 
prediction quality and prediction speed, as scoring many candidates can be prohibitively slow.

The current state-of-the-art focuses on using BERT models for pre-training \citep{devlin-etal-2019-bert}, which employ large text corpora on general subjects: Wikipedia and the Toronto Books Corpus \citep{Zhu2015AligningBA}. 
Two classes of fine-tuned architecture are typically built on top: 
Bi-encoders and Cross-encoders.
 Cross-encoders \citep{wolf2019transfertransfo,vig2019comparison}, which perform full (cross) self-attention over a given input and label candidate, tend to attain much higher accuracies than their counterparts, Bi-encoders \citep{training_millions,dinan2019wizard}, which perform self-attention over the input and candidate label separately and combine them at the end for a final representation. As the representations are separate, Bi-encoders 
are able to cache the encoded candidates, and reuse these representations for each input resulting in fast prediction times.
 Cross-encoders must recompute the encoding for each input and label; as a result, they are prohibitively slow at test time.

In this work, we provide novel contributions that improve both the quality and speed axes over the current  state-of-the-art.
%
    We introduce the Poly-encoder, an architecture with an additional learnt attention mechanism that represents more global features from which to perform self-attention, resulting in performance gains over Bi-encoders and large speed gains over Cross-Encoders.
%
To pre-train our architectures, we show that choosing abundant data more similar to our downstream task also brings significant gains over BERT pre-training.  This is true across all different architecture choices and downstream tasks we try.

We conduct experiments comparing the new approaches, in addition to analysis of what works best for various setups of existing methods, 
on four existing datasets in the domains of dialogue and 
information retrieval (IR),
with pre-training strategies based on  Reddit \citep{training_millions} compared to Wikipedia/Toronto Books (i.e., BERT). 
We obtain a new state-of-the-art  on all four datasets with our best architectures and pre-training strategies, as well as providing practical implementations for real-time use.
Our code and models will be released open-source. 


\section{Related Work}

The task of scoring candidate labels given an input context is a classical
problem in machine learning. 
While multi-class classification is a special case,
the more general task involves candidates as structured objects
rather than discrete classes; in this work we consider the inputs and the candidate labels to be sequences of text. 

There is a broad class of models that map the input and a candidate label separately into a common feature space wherein typically a dot product, cosine or (parameterized) non-linearity is used to measure their similarity.  
We refer to these models as  {\em Bi-encoders}.
Such methods include vector space models \citep{salton1975vector},
LSI \citep{deerwester1990indexing},
supervised embeddings \citep{bai2009supervised,wu2018starspace} and classical siamese networks \citep{bromley1994signature}.
For the next utterance prediction tasks we consider in this work, several Bi-encoder neural approaches have been considered, in particular Memory Networks \citep{zhang2018personalizing} and Transformer Memory networks \citep{dinan2019wizard}
as well as LSTMs \citep{lowe2015ubuntu} and CNNs \citep{kadlec2015improved} which encode input and 
candidate label separately.
A major advantage of Bi-encoder methods 
is their ability to cache the representations of a large, fixed candidate set. Since the candidate encodings are independent of the input, Bi-encoders are very efficient during evaluation.

 Researchers have also studied a more rich class of models we refer to as {\em Cross-encoders}, which make no assumptions on the similarity scoring function between input and candidate label. Instead, the concatenation of the input and a candidate serve as a new input to a nonlinear function that scores their match based on any dependencies it wants. This has been explored with Sequential Matching Network CNN-based architectures \citep{wu2016sequential},  Deep Matching Networks \citep{yang2018response},
Gated Self-Attention \citep{zhang2018modeling},
and most recently transformers \citep{wolf2019transfertransfo,vig2019comparison,urbanek2019learning}.
For the latter, concatenating the two sequences of  text results in applying self-attention  at every layer. This yields rich interactions between the input context and the candidate, as every word in the candidate label can attend to every word in the input context, and vice-versa.
\citet{urbanek2019learning} employed
pre-trained BERT models, and fine-tuned both Bi- and  Cross-encoders, explicitly comparing them on dialogue and action tasks,
and finding that Cross-encoders perform better.
However, the performance gains come at a steep computational cost. Cross-encoder representations are much slower to compute, rendering some applications infeasible.

\section{Tasks} \label{sec:tasks}
We consider the tasks of sentence selection in dialogue and article search in IR.
The former is
a task extensively studied and recently featured in two competitions: the Neurips ConvAI2 competition \citep{dinan2019second},
and the DSTC7 challenge, Track 1 \citep{yoshino2019dialog,arxiv18disentangle,dstc19task1}. We compare on those two tasks and in addition, we also test on the popular Ubuntu V2 corpus \citep{lowe2015ubuntu}. For IR, we use the Wikipedia Article Search task of \citet{wu2018starspace}.


The ConvAI2 task is based on the Persona-Chat dataset \citep{zhang2018personalizing}
which involves dialogues between pairs of speakers. Each speaker is given a persona, which is a few sentences that describe a character they will imitate, e.g. ``I love romantic movies'', and is instructed to get to know the other.
Models should then condition their chosen response on the dialogue history and the lines of persona.
As an automatic metric in the competition,
for each response, the model has to pick the 
correct annotated utterance from a set of 20 choices, where the remaining 19 were other randomly chosen utterances from the evaluation set. 
Note that in a final system however, one would retrieve from the entire training set of over 100k utterances, but this is avoided for speed reasons in common evaluation setups.
The best performing competitor out of 23 entrants in this task
achieved 80.7\% accuracy on the test set utilizing a pre-trained Transformer 
fine-tuned for this task \citep{wolf2019transfertransfo}.

The DSTC7 challenge (Track 1) consists of
 conversations extracted from Ubuntu chat logs, where one partner receives technical support for various Ubuntu-related problems from the other. The best performing competitor (with 20 entrants in Track 1) in this task achieved 64.5\% R@1 \citep{chen_noetic}. 
Ubuntu V2 is a similar but larger popular corpus, created before the competition \citep{lowe2015ubuntu}; we report results for this dataset as well, as there are many existing results on it. 

Finally, we evaluate on Wikipedia Article Search \citep{wu2018starspace}. Using the 2016-12-21 dump of English Wikipedia ($\sim$5M articles),  the task is given a sentence from an  article as a search query, find the article it came
from. Evaluation ranks the true article (minus the sentence) against 10,000 other  articles using retrieval metrics. This mimics a web search like scenario where one would like to search for the most relevant articles (web documents). The best reported method is the learning-to-rank embedding model, StarSpace, which outperforms fastText, SVMs, and other baselines.

We summarize all four datasets and their statistics in Table \ref{table:datasets}.


\begin{table}[h]
\center
\begin{small}
\begin{tabular}{|c|cccc|}
\hline
             &  ConvAI2 &  DTSC7 & Ubuntu V2 & Wiki Article Search\\
\hline
Train Ex. & 131,438 & 100,000 &  1,000,000 &  5,035,182\\
Valid Ex. & 7,801&  10,000 & 19,560        &  9,921 \\
Test  Ex. & 6634 & 5,000 &  18,920         &  9,925\\
Eval Cands per Ex. &  20  & 100 &  10      &  10,001\\
\hline
\end{tabular}
\end{small}
\caption{Datasets used in this paper.}
\label{table:datasets}
\vspace{-0.5em}
\end{table}

\section{Methods}

In this section we describe the various models and methods that we explored.

\subsection{Transformers and Pre-training Strategies}
\paragraph{Transformers}
Our Bi-, Cross-, and Poly-encoders, described in sections \ref{subsection:Bi_Encoders}, \ref{subsection:Cross_Encoder} and \ref{subsection:Poly_Encoder} respectively, are based on large pre-trained transformer models with the same architecture and dimension as BERT-base \citep{devlin-etal-2019-bert}, which has 12 layers, 12 attention heads, and a hidden size of 768. As well as considering the BERT pre-trained weights, we also explore our own pre-training schemes. Specifically, we pre-train two more transformers from scratch using the exact same architecture as BERT-base. 
One uses a similar training
setup as in BERT-base, training on 150 million of examples of [INPUT, LABEL] extracted from Wikipedia and the Toronto Books Corpus,  while the other is trained on  174 million  examples of [INPUT, LABEL] extracted from the online platform Reddit \citep{training_millions}, which is a dataset more adapted to dialogue. 
The former is performed to verify that reproducing a BERT-like setting gives us the same results as reported previously, while the latter tests whether pre-training on data
more similar to the downstream tasks of interest helps.
For training both new setups we used  XLM  \citep{lample2019cross}.

\paragraph{Input Representation}

Our pre-training input is the concatenation of input and label [INPUT,LABEL], 
where both are surrounded with the special token [S], 
following \citet{lample2019cross}. 
When pre-training on Reddit, the input is the context,
and the label is the next utterance. When pre-training on Wikipedia and Toronto Books, as in \citet{devlin-etal-2019-bert}, the input is one sentence and the label the next sentence in the text. 
Each input token is represented as the sum of three embeddings: the token embedding, the position (in the sequence) embedding and the segment embedding.
Segments for input tokens are 0, and for label tokens are 1.

\paragraph{Pre-training Procedure}
\label{pretraining}
Our pre-training strategy involves training with a masked language model (MLM) task identical to the one in \citet{devlin-etal-2019-bert}. In the pre-training on Wikipedia and Toronto Books we add a next-sentence prediction task identical to BERT training. In the pre-training on Reddit, we add a next-utterance prediction task, which is slightly different from the previous one as an utterance can be composed of several sentences. During training 50\% of the time the candidate is the actual next sentence/utterance and 50\% of the time it is a sentence/utterance randomly taken from the dataset. We alternate between batches of the MLM task and the next-sentence/next-utterance prediction task.
Like in \citet{lample2019cross} we use the Adam optimizer with learning rate of 2e-4, $\beta_1 = 0.9$,
$\beta_2 = 0.98$, no L2 weight decay, linear learning
rate warmup, and inverse square root
decay of the learning rate. We use a dropout probability of 0.1 on all layers, and a batch of 32000 tokens composed of concatenations [INPUT, LABEL] with similar lengths. We train the model on 32 GPUs for 14 days.

\paragraph{Fine-tuning}
After pre-training, one can then fine-tune for the multi-sentence selection task of choice, in our case one of the four tasks 
from Section~\ref{sec:tasks}.
We consider three architectures with which we fine-tune the transformer:
the Bi-encoder, Cross-encoder and newly proposed Poly-encoder.

\subsection{Bi-encoder}
\label{subsection:Bi_Encoders}
In a Bi-encoder,
both the input context and the candidate label are encoded into vectors:
$$
\begin{array}{l}
        y_{ctxt} = red(T_1(ctxt)) ~~~~~~~~~ y_{cand} = red(T_2(cand)) \\
\end{array}
$$
 where $T_1$ and $T_2$ are two transformers that have been pre-trained following the procedure described in \ref{pretraining}; they initially start with the same weights, but are allowed to update separately during fine-tuning. $T(x)=h_1,..,h_N$ is the output of a transformer T
 and $red(\cdot)$ is a function that reduces that sequence of vectors  into one vector. 
 As the input and the label are encoded separately, segment tokens are 0 for both.
 To resemble what is done during our pre-training, both the input and label are surrounded by the special token [S] and therefore $h_1$ corresponds to [S].
 
 We considered three ways of reducing the output into one representation via $red(\cdot)$: choose the first output of the transformer (corresponding to the special token [S]),  compute the average over all outputs or the average over the first $m \le N$ outputs. We compare them in Table \ref{table:reduction} in the Appendix. We use the first output of the transformer in our experiments as it gives slightly better results.

\paragraph{Scoring} The score of a candidate $cand_i$ is given by the dot-product $s(ctxt, cand_i)=y_{ctxt} \cdot y_{cand_i}$. 
%
%
The network is trained to minimize a cross-entropy loss in which the logits are $y_{ctxt} \cdot y_{cand_1},...,  y_{ctxt} \cdot y_{cand_n}$, where $cand_1$ is the correct label and the others are chosen from the training set. Similar to  \citet{training_millions}, during training we consider the other labels in
the batch as negatives. This allows for much faster training, as we can reuse the embeddings computed for each candidate, and also use a larger batch size; e.g., in our experiments on ConvAI2, we were able to use batches of 512 elements.

\paragraph{Inference speed} In the setting of retrieval over known candidates, a Bi-encoder allows for the precomputation of the embeddings of all possible candidates of the system. After the context embedding $y_{ctxt}$ is computed, the only operation remaining is a dot product between $y_{ctxt}$ and every candidate embedding, which can scale to millions of candidates on a modern GPU, and potentially billions using nearest-neighbor libraries such as FAISS \citep{JDH17}.


\subsection{Cross-encoder}
\label{subsection:Cross_Encoder}
The Cross-encoder allows for rich interactions between the input context and candidate label, as they are jointly encoded to obtain a final representation. 
Similar to the procedure in pre-training, the context and candidate are surrounded by the special token [S] and concatenated into a single vector, which is encoded using one transformer. We consider the first output of the transformer as the context-candidate embedding:
$$
\begin{array}{l}
        y_{ctxt,cand} = h_1 = first(T(ctxt,cand)) \\
\end{array}
$$
where $first$ is the  function that takes the first vector of the sequence of vectors produced by the transformer. By using a single transformer, the Cross-encoder is able to perform self-attention between the context and candidate, resulting in a richer extraction mechanism than the Bi-encoder. As the candidate label can attend to the input context during the layers of the transformer, the Cross-encoder can produce a candidate-sensitive input representation, which the Bi-encoder cannot. For example, this allows it to select useful input features per candidate. 

\paragraph{Scoring}
To score one candidate, a linear layer $W$ is applied to the embedding $y_{ctxt,cand}$ to reduce it from a vector to a scalar: 
$$
\begin{array}{l}
        s(ctxt,cand_i) = y_{ctxt,cand_i}W\\
\end{array}
$$

Similarly to what is done for the Bi-encoder, the network is trained to minimize a cross entropy loss where the logits are $s(ctxt,cand_1), ..., s(ctxt,cand_n)$, where $cand_1$ is the correct candidate and the others are negatives taken from the training set. Unlike in the Bi-encoder, we cannot recycle the other labels of the batch as negatives, so we use external negatives provided in the training set. The Cross-encoder uses much more memory than the Bi-encoder, resulting in a much smaller batch size.

\paragraph{Inference speed}

Unfortunately, the Cross-encoder does not allow for precomputation of the candidate embeddings. At inference time, every candidate must be concatenated with the input context and must go through a forward pass of the entire model. Thus, this method cannot scale to a large amount of candidates. We discuss this bottleneck further in Section \ref{section:timing}.
\begin{figure}[t]
    \center
  \includegraphics[width=0.8\textwidth]{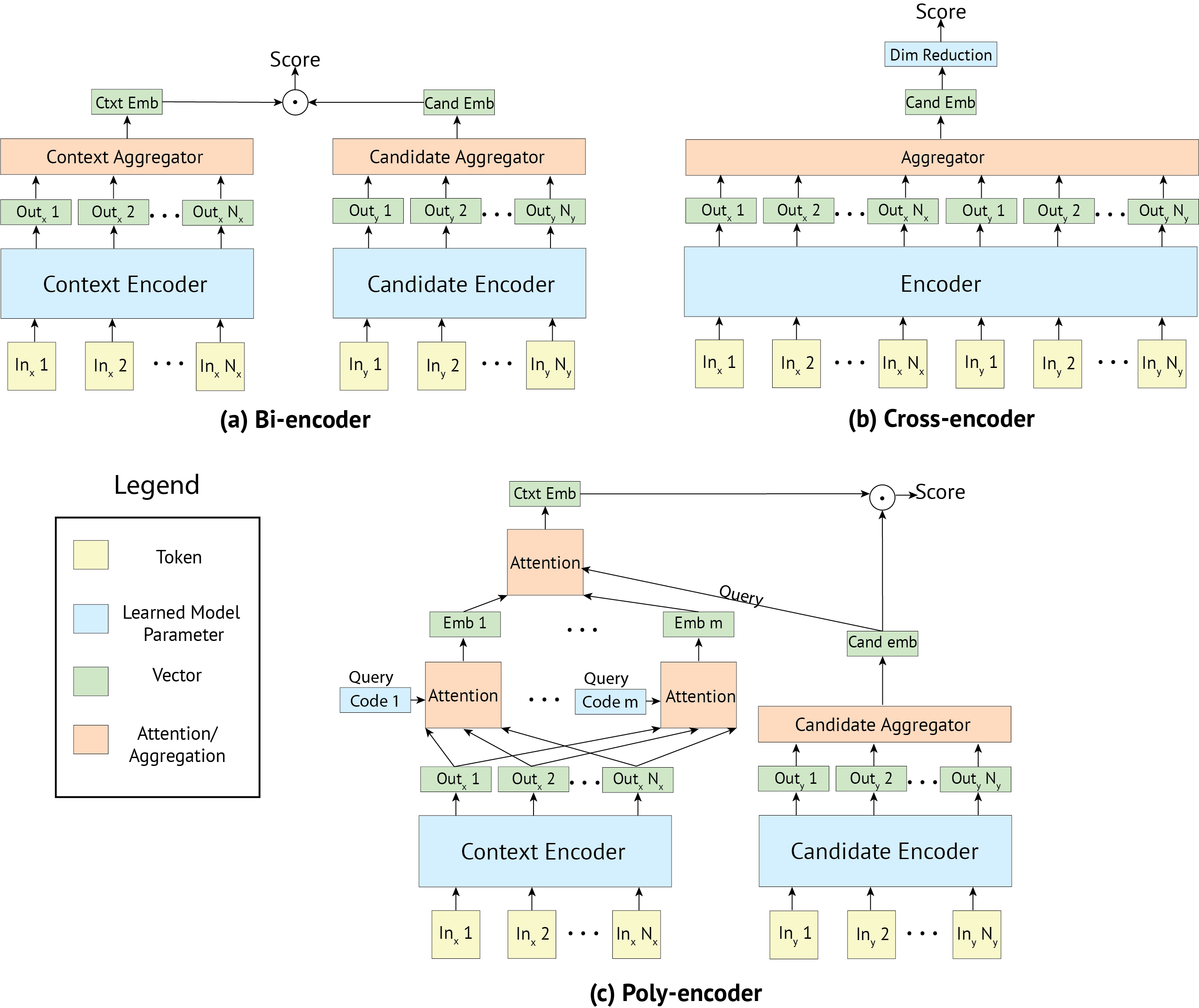}
  \caption{Diagrams of the three model architectures we consider. (a) The Bi-encoder encodes the context and candidate separately, allowing for the caching of candidate representations during inference. (b) The Cross-encoder jointly encodes the context and candidate in a single transformer, yielding richer interactions between context and candidate at the cost of slower computation. (c) The Poly-encoder combines the strengths of the Bi-encoder and Cross-encoder by both allowing for caching of candidate representations and adding a final attention mechanism between global features of the input and a given candidate to give richer interactions
  before computing a final score.}
  \label{architectures}
\end{figure}

\subsection{Poly-encoder}

\label{subsection:Poly_Encoder}
The Poly-encoder architecture aims to get the best of both worlds from the Bi- and  Cross-encoder. A given candidate label is represented by one vector as in the Bi-encoder, which allows for caching candidates for fast inference time, while the input context is jointly encoded with the candidate, as in the Cross-encoder, allowing the extraction of more information.

The Poly-encoder uses two separate transformers for the context and label like a Bi-encoder, and the candidate is encoded into a single vector $y_{cand_i}$. As such, the Poly-encoder method can be implemented using a precomputed cache of encoded responses. 
However, the input context, which is typically much longer than a candidate, is represented with $m$ vectors ($y^1_{ctxt}..y^m_{ctxt}$) instead of just one as in the Bi-encoder, where $m$ will influence the inference speed.  
To obtain these $m$ global features that represent the input, we learn $m$ context codes $(c_1, ..., c_m)$, where $c_i$ extracts representation $y^i_{ctxt}$ by attending over all the outputs of the previous layer. That is, we obtain $y^i_{ctxt}$ using: 
$$ y^i_{ctxt} = \sum_{j}{w^{c_i}_j h_j} \textrm{\hspace{5 mm} where \hspace{5 mm}} (w^{c_i}_1, .., w^{c_i}_N) = \textrm{softmax}(c_i \cdot h_1, .. , c_i \cdot h_N)$$
The $m$ context codes are randomly initialized, and learnt during finetuning. 

Finally, given our  $m$ global context features, we attend over them using $y_{cand_i}$ as the query:
$$ y_{ctxt} = \sum_{i}{w_i y^i_{ctxt}} \textrm{\hspace{5 mm} where \hspace{5 mm}}(w_1, .., w_m) = \textrm{softmax}(y_{cand_i} \cdot y_{ctxt}^1, .. , y_{cand_i} \cdot y_{ctxt}^m)$$
The final score for that candidate label is then $y_{ctxt} \cdot y_{cand_i}$ as in a Bi-encoder.
As $m < N$, where $N$ is the number of tokens, and the context-candidate attention is only performed at the top layer, this is far faster than the Cross-encoder's full self-attention.

\section{Experiments}
We perform a variety of experiments to test our model architectures and training strategies over four tasks. For metrics, we measure Recall@$k$ where each test example has $C$ possible candidates to select from, 
abbreviated to R@$k$/$C$, as well as mean reciprocal rank (MRR).

\subsection{Bi-encoders and Cross-encoders}
\label{subsection:biencoder}
 
We first investigate fine-tuning the Bi- and Cross-encoder architectures initialized with the weights provided by \citet{devlin-etal-2019-bert}, studying the choice of other hyperparameters (we explore our own pre-training schemes in section \ref{section:domainspepretraining}). In the case of the Bi-encoder, we can use a large number of negatives by considering the other batch elements as negative training samples, avoiding recomputation of their embeddings. On 8 Nvidia Volta v100 GPUs and using half-precision operations (i.e. float16 operations), we can reach batches of 512 elements on ConvAI2. Table \ref{table:negs_bi} shows that in this setting, we obtain higher performance with a larger batch size, i.e. more negatives, where 511 negatives yields the best results. For the other tasks, we keep the batch size at 256, as the longer sequences in those datasets uses more memory.
The Cross-encoder is more computationally intensive, as the embeddings for the (context, candidate) pair must be recomputed each time. We thus limit its batch size to 16 and provide negatives random samples from the training set. For DSTC7 and Ubuntu V2, we choose 15 such negatives; For ConvAI2, the dataset provides 19 negatives.
\begin{table}[h]
\center
\begin{tabular}{|c|c|c|c|c|c|c|}
\hline
Negatives & 31 & 63 & 127 & 255 & 511\\
\hline
R@1/20 & 81.0 & 81.7 & 82.3 & 83.0 & \textbf{83.3} \\

\hline
\end{tabular}
\caption{Validation performance  on ConvAI2 after fine-tuning a Bi-encoder pre-trained with BERT, averaged over 5 runs. The batch size is the number of training negatives + 1 as we use the other elements of the batch as negatives during training. 
}
\label{table:negs_bi}
\vspace{-0.5em}
\end{table}

The above results are reported with Bi-encoder aggregation based on the first output. Choosing the average over all outputs instead is very similar but slightly worse (83.1, averaged over 5 runs).
We also tried to add further non-linearities instead of the inner product of the two representations, but could not obtain improved results over the simpler architecture (results not shown).



We tried two optimizers: Adam \citep{kingma2014adam} with weight decay of 0.01 (as recommended by \citep{devlin-etal-2019-bert}) and Adamax  \citep{kingma2014adam} without weight decay; based on validation set performance, we choose to fine-tune with Adam when using the BERT weights.
The learning rate is initialized to 5e-5 with a warmup of 100 iterations for Bi- and Poly-encoders, and 1000 iterations for the Cross-encoder. The learning rate decays by a factor of 0.4 upon plateau of the loss evaluated on the valid set every half epoch. In Table \ref{table:frozen_params} we show validation performance when fine-tuning various layers of the weights provided by \citep{devlin-etal-2019-bert}, using Adam with decay optimizer.  Fine-tuning the entire network is important, with the exception of the word embeddings.
\begin{table}[h]
\center
\setlength\tabcolsep{3.2pt} 
\begin{tabular}{|c|c|c|}
\hline
Fine-tuned parameters & Bi-encoder & Cross-encoder \\
\hline

Top layer                  & 74.2 & 80.6 \\
Top 4 layers            & 82.0 & 86.3 \\
All but Embeddings  & \textbf{83.3} & \textbf{87.3}  \\
Every Layer & 83.0 & 86.6 \\

\hline
\end{tabular}
\caption{Validation performance (R@1/20) on ConvAI2 using pre-trained weights of  BERT-base with different  parameters fine-tuned. Average over 5 runs (Bi-encoders) or 3 runs (Cross-encoders).}
\label{table:frozen_params}
\vspace{-1.0em}
\end{table}

With the setups described above, we fine-tune the Bi- and Cross-encoders on the datasets,
and report the results in Table \ref{table:best_scores}. On the first three tasks, our Bi-encoders and Cross-encoders outperform the best existing approaches in the literature when we fine-tune from BERT weights. E.g., the Bi-encoder reaches 81.7\% R@1 on ConvAI2 and 66.8\% R@1 on DSTC7, while the Cross-encoder achieves higher scores of 84.8\% R@1 on ConvAI2 and 67.4\% R@1 on DSTC7.
Overall, Cross-encoders outperform all previous approaches on the three dialogue tasks, including our Bi-encoders (as expected). We do not report fine-tuning of BERT for Wikipedia IR as we cannot guarantee the test
set is not part of the pre-training for that dataset. In addition, Cross-encoders are also too slow to evaluate on the evaluation setup of that task, which has 10k candidates.

\begin{table*}[h]
\center
\setlength\tabcolsep{3.2pt} 
\small
\begin{tabular}{|l|c|c|c|c|c|c|c|}
\hline
 Dataset & \multicolumn{1}{c|}{ConvAI2} & \multicolumn{2}{c|}{DSTC 7} & \multicolumn{2}{c|}{Ubuntu v2} & Wikipedia IR\\
 \hline
 split  & test  & \multicolumn{2}{c|}{test}  & \multicolumn{2}{c|}{test} & test \\
 \hline
metric & R@1/20 & R@1/100 & MRR & R@1/10  & MRR & R@1/10001\\
\hline
\hline
\citep{wolf2019transfertransfo} & 80.7 & & & & & \\
\hline
\citep{gu} & - & 60.8 & 69.1 & -  & - & - \\
\hline
\citep{chen_noetic} & - & 64.5 &  73.5 & -  & -  & -  \\
\hline
\citep{yoon} & -  & -  & -  &  65.2 & - & -\\
\hline
\citep{dong_et_hal_ubuntu2}  & -  & -  & -  &  75.9 & 84.8 & -  \\
\hline
\citep{wu2018starspace}  & -  & -  & -  &  - & - & 56.8  \\
\hline
\hline
 \multicolumn{7}{|l|}{pre-trained BERT weights from \citep{devlin-etal-2019-bert} - Toronto Books + Wikipedia}  \\
\hline

Bi-encoder & 81.7 $\pm$ 0.2 & 66.8 $\pm$ 0.7 & 74.6 $\pm$ 0.5 & 80.6 $\pm$ 0.4 & 88.0 $\pm$ 0.3 & -  \\ \hline 





Poly-encoder 16 & 83.2 $\pm$ 0.1 & 67.8 $\pm$ 0.3 & 75.1 $\pm$ 0.2 & 81.2 $\pm$ 0.2 & 88.3 $\pm$ 0.1 & -  \\ \hline 

Poly-encoder 64 & 83.7 $\pm$ 0.2 & 67.0 $\pm$ 0.9 & 74.7 $\pm$ 0.6 & 81.3 $\pm$ 0.2 & 88.4 $\pm$ 0.1 & -  \\ \hline 

Poly-encoder 360 & 83.7 $\pm$ 0.2 & 68.9 $\pm$ 0.4 & 76.2 $\pm$ 0.2 & 80.9 $\pm$ 0.0 & 88.1 $\pm$ 0.1 & -  \\ \hline 

Cross-encoder & 84.8 $\pm$ 0.3 & 67.4 $\pm$ 0.7  & 75.6 $\pm$ 0.4 & 82.8 $\pm$ 0.3 & 89.4 $\pm$ 0.2 & -  \\
\hline 
\hline 
 \multicolumn{7}{|l|}{Our pre-training on Toronto Books + Wikipedia}  \\ 
 \hline 
Bi-encoder & 82.0 $\pm$ 0.1 & 64.5 $\pm$ 0.5 & 72.6 $\pm$ 0.4 & 80.8 $\pm$ 0.5 & 88.2 $\pm$ 0.4 & -  \\ \hline 




Poly-encoder 16 & 82.7 $\pm$ 0.1 & 65.3 $\pm$ 0.9 & 73.2 $\pm$ 0.7 & 83.4 $\pm$ 0.2 & 89.9 $\pm$ 0.1 & -  \\ \hline 

Poly-encoder 64 & 83.3 $\pm$ 0.1 & 65.8 $\pm$ 0.7 & 73.5 $\pm$ 0.5 & 83.4 $\pm$ 0.1 & 89.9 $\pm$ 0.0 & -  \\ \hline 

Poly-encoder 360 & 83.8 $\pm$ 0.1 & 65.8 $\pm$ 0.7 & 73.6 $\pm$ 0.6 &83.7 $\pm$ 0.0 & 90.1 $\pm$ 0.0 & -  \\ \hline 

Cross-encoder & 84.9 $\pm$ 0.3 & 65.3 $\pm$ 1.0 & 73.8 $\pm$ 0.6 & 83.1 $\pm$ 0.7 & 89.7 $\pm$ 0.5 & -  \\ 
\hline 
\hline 
\multicolumn{7}{|l|}{Our pre-training on Reddit}  \\ 
\hline 
Bi-encoder & 84.8 $\pm$ 0.1 & 70.9 $\pm$ 0.5 & 78.1 $\pm$ 0.3 & 83.6 $\pm$ 0.7 & 90.1 $\pm$ 0.4 & 71.0  \\ \hline 




Poly-encoder 16 & 86.3 $\pm$ 0.3 & 71.6 $\pm$ 0.6 & 78.4 $\pm$ 0.4 & 86.0 $\pm$ 0.1 & 91.5 $\pm$ 0.1 & 71.5 \\ \hline 

Poly-encoder 64 & 86.5 $\pm$ 0.2 & 71.2 $\pm$ 0.8 & 78.2 $\pm$ 0.7 & 85.9 $\pm$ 0.1 & 91.5 $\pm$ 0.1 & 71.3\\ \hline 

Poly-encoder 360 & 86.8 $\pm$ 0.1 & 71.4 $\pm$ 1.0 & 78.3 $\pm$ 0.7 & 85.9 $\pm$ 0.1 & 91.5 $\pm$ 0.0 & \textbf{71.8}\\ \hline 

Cross-encoder & \textbf{87.9 $\pm$ 0.2} & \textbf{71.7 $\pm$ 0.3} & \textbf{79.0 $\pm$ 0.2} & \textbf{86.5 $\pm$ 0.1} & \textbf{91.9 $\pm$ 0.0} & -  \\ \hline 

\end{tabular}
\caption{Test performance of Bi-, Poly- and Cross-encoders on
our selected tasks
.}
\label{table:best_scores}
\vspace{-0.5em}
\end{table*}

\subsection{Poly-encoders}
\label{subsection:poly}

We train the Poly-encoder using the same batch sizes and optimizer choices as in the Bi-encoder experiments. Results are reported in Table \ref{table:best_scores} for various values of $m$ context vectors.


The Poly-encoder outperforms the Bi-encoder on all the tasks, with more codes generally yielding larger improvements.  Our recommendation is thus to use as large a code size as compute time allows
(see Sec. \ref{section:timing}).
On DSTC7, the Poly-encoder architecture with BERT pretraining
reaches 68.9\% R1 with 360 intermediate context codes; this actually outperforms the Cross-encoder result (67.4\%) and is noticeably better than our Bi-encoder result (66.8\%).
Similar conclusions are found on Ubuntu V2 and ConvAI2, although in the latter Cross-encoders
give slightly better results.


We note that since reporting our results, the authors of \citet{li2019acute} have conducted a human evaluation study on ConvAI2, in which our Poly-encoder architecture outperformed all other models compared against, both generative and retrieval based, including the winners of the competition.

\subsection{Domain-specific Pre-training}
\label{section:domainspepretraining}

We fine-tune our Reddit-pre-trained transformer on all four tasks; we additionally fine-tune a transformer that was pre-trained on the same datasets as BERT, specifically Toronto Books + Wikipedia. 
When using our pre-trained weights, we use the Adamax optimizer and optimize all the layers of the transformer including the embeddings. As we do not use weight decay, the weights of the final layer are much larger than those in the final layer of BERT; to avoid saturation of the attention layer in the Poly-encoder, we re-scaled the last linear layer so that the standard deviation of its output matched that of BERT, which we found necessary to achieve good results.
We report results of fine-tuning with our pre-trained weights in Table \ref{table:best_scores}. We show that pre-training on Reddit gives further state-of-the-art performance over our previous results with BERT, a finding that we see for {all} three dialogue tasks, and {all} three architectures. 

The results obtained with fine-tuning on our own transformers pre-trained on Toronto Books + Wikipedia are very similar to those obtained with the original BERT weights, indicating that the choice of dataset used to pre-train the models impacts the final results, not some other detail in our training.
 Indeed, as the two settings pre-train with datasets of similar size, we can conclude that choosing a pre-training task (e.g. dialogue data) that is similar to the downstream tasks of interest (e.g. dialogue) is a likely explanation for these performance gains, in line with previous results showing multi-tasking with similar tasks is more useful than with dissimilar ones \citep{caruana1997multitask}.

\subsection{Inference Speed}
\label{section:timing}

An important motivation for the Poly-encoder architecture is to achieve better results than the Bi-encoder while also performing at a reasonable speed. Though the Cross-encoder generally yields strong results, it is prohibitively slow. We perform speed experiments to determine the trade-off of improved performance from the Poly-encoder. Specifically, we predict the next utterance for 100 dialogue examples in the ConvAI2 validation set, where the model scores $C$ candidates (in this case, chosen from the training set). We perform these experiments on both CPU-only and  GPU setups. CPU computations were run on an 80 core Intel Xeon processor CPU E5-2698. GPU computations were run on a single  Nvidia Quadro GP100 using cuda 10.0 and cudnn 7.4.

We show the average time per example for each architecture 
in Table \ref{timing}. The difference in timing between the 
Bi-encoder and the Poly-encoder architectures is rather minimal when there are only 1000 candidates for the model to consider. The difference is more pronounced when considering 100k candidates, a more realistic setup, as we see a 5-6x slowdown for the Poly-encoder variants. Nevertheless, both models are still tractable. 
The Cross-encoder, however, is 2 orders of magnitude slower than the
Bi-encoder and Poly-encoder, rendering it intractable for real-time inference, e.g. when interacting with a dialogue agent, or retrieving from a large set of documents.   
Thus,  Poly-encoders, given their desirable performance and speed trade-off, are the preferred method.

We additionally report training times in the Appendix, Table \ref{table:training_times}. Poly-encoders also have the benefit of being
3-4x faster to train than Cross-encoders (and are similar in training time to Bi-encoders).

\begin{table}
\center
\setlength\tabcolsep{4pt}
\begin{tabular}{|c|c|c|c|c|}
\hline
 & \multicolumn{4}{c|}{Scoring time (ms)}\\
\hline
 & \multicolumn{2}{c|}{CPU} & \multicolumn{2}{c|}{GPU}\\
\hline
Candidates & 1k  & 100k& 1k  & 100k \\
\hline
\hline
Bi-encoder & 115 & 160 & 19 & 22\\
\hline
Poly-encoder 16 & 122 & 678 & 18 & 38 \\
\hline
Poly-encoder 64 & 126 & 692 & 23 & 46 \\
\hline
Poly-encoder 360 & 160 & 837 & 57 & 88 \\
\hline
Cross-encoder & 21.7k & 2.2M* & 2.6k & 266k*\\

\hline

\end{tabular}
\caption{Average time in milliseconds to predict the next dialogue utterance from $C$ possible candidates on ConvAI2. * are inferred.}
\label{timing}
\vspace{-1.25em}
\end{table}

\section{Conclusion}
In this paper we present new architectures and pre-training strategies
for deep bidirectional transformers in candidate selection tasks.
We introduced the Poly-encoder method, which provides a mechanism for attending over the context using the label candidate, while maintaining the ability to precompute each candidate's representation, which allows for fast real-time inference in a production setup, giving an improved 
trade off between accuracy and speed. We provided an 
experimental analysis of those trade-offs for Bi-, 
Poly- and
Cross-encoders, showing that Poly-encoders are more accurate than Bi-encoders, 
while being far faster than Cross-encoders, which are impractical for real-time use.
In terms of training these architectures,
we showed that pre-training strategies more closely related to the downstream task bring strong improvements. In particular, 
pre-training from scratch on Reddit allows us to outperform the results we obtain with BERT, 
a result that holds for all three model architectures and all three dialogue datasets we tried.
However, the methods introduced in this work 
are not specific to dialogue, and can be used for any task
where one is scoring a set of candidates, which we showed  
for an information retrieval task as well. 


\bibliography{iclr2020_conference}
\bibliographystyle{iclr2020_conference}
\newpage
\appendix
\section{Training Time}

We report the training time on 8 GPU Volta 100 for the 3 datasets considered and for 4 types of models in Table \ref{table:training_times}.

\begin{table}[h]
\center
\begin{small}
\begin{tabular}{|c|c|c|c|}
\hline
Dataset & ConvAI2 & DSTC7 & UbuntuV2\\
\hline
Bi-encoder & 2.0 & 4.9 & 7.9 \\
\hline
Poly-encoder 16 & 2.7 & 5.5 & 8.0 \\
\hline
Poly-encoder 64 & 2.8 & 5.7 & 8.0 \\
\hline
Cross-encoder64 & 9.4 & 13.5 & 39.9 \\
\hline

\end{tabular}
\end{small}
\caption{Training time in hours.}
\label{table:training_times}
\end{table}

\section{Reduction layer in Bi-encoder}

We provide in Table \ref{table:reduction} the results obtained for different types of reductions on top of the Bi-encoder. Specifically we compare the Recall@1/20 on the ConvAI2 validation set when taking the first output of BERT, the average of the first 16 outputs, the average of the first 64 outputs and all of them except the first one ([S]).

\begin{table}[h]
\center
\begin{small}
\begin{tabular}{|c|c|}
\hline
Setup & ConvAI2 valid Recall@1/20\\
\hline
First output & 83.3 \\
\hline
Avg first 16 outputs & 82.9 \\
\hline
Avg first 64 outputs & 82.7 \\
\hline
Avg all outputs & 83.1\\
\hline

\end{tabular}
\end{small}
\caption{Bi-encoder results on the ConvAI2 valid set for different choices of function $red(\cdot)$.}
\label{table:reduction}
\end{table}

\section{Alternative Choices for Context Vectors}
\label{section:alternative_ctxt_vecs}

\begin{figure*}[h]
\centering
  \includegraphics[width=0.9\textwidth]{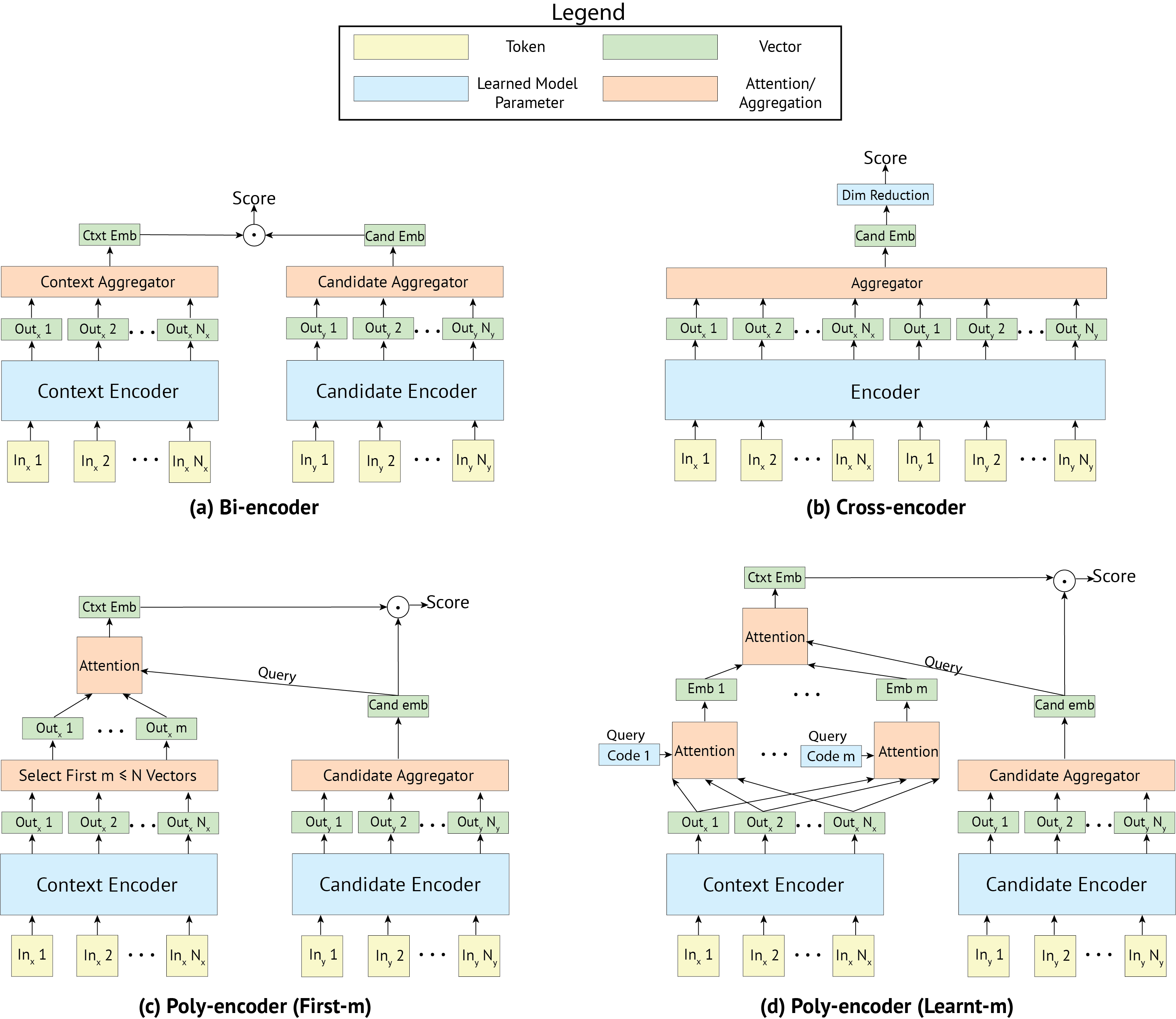}
  \caption{(a) The Bi-encoder (b) The Cross-encoder (c) The Poly-encoder with first $m$ vectors. (d) The Poly-encoder with $m$ learnt codes.}
  \label{figure_all_polys}
\end{figure*}

We considered a few other ways to derive the context vectors $(y^1_{ctxt}, ..., y^m_{ctxt})$ of the Poly-encoder from the output $(h^1_{ctxt}, ..., h^N_{ctxt})$ of the underlying transformer: 
\begin{itemize}
    \item  Learn $m$ codes $(c_1, ..., c_m)$, where $c_i$ extracts representation $y^i_{ctxt}$ by attending over all the outputs  $(h^1_{ctxt}, ..., h^N_{ctxt})$. This method is denoted ``Poly-encoder (Learnt-codes)" or ``Poly-encoder (Learnt-m)", and is the method described in section \ref{subsection:Poly_Encoder}
    \item Consider the first $m$ outputs $(h^1_{ctxt}, ..., h^m_{ctxt})$. This method is denoted ``Poly-encoder (First $m$ outputs)" or ``Poly-encoder (First-m)". Note that when $N < m$, only $m$ vectors are considered.
    \item Consider the last $m$ outputs.
    \item Consider the last $m$ outputs concatenated with the first one, $h^1_{ctxt}$ which plays a particular role in BERT as it corresponds to the special token [S].
\end{itemize}
The performance of those four methods is evaluated on the validation set of Convai2 and DSTC7 and reported on Table \ref{table:all_poly_types}. The first two methods are shown in Figure \ref{figure_all_polys}. We additionally provide the inference time for a given number of candidates coming from the Convai2 dataset on Table \ref{all_timing}.

\begin{table*}[h]
\center
\begin{tabular}{|l|c|c|c|c|}
\hline
 Dataset & \multicolumn{2}{c|}{ConvAI2} & \multicolumn{2}{c|}{DSTC 7}\\ 
 \hline
 split & dev & test & dev & test \\
 \hline
metric & R@1/20 & R@1/20 & R@1/100 & R@1/100 \\
\hline
\hline
\citep{wolf2019transfertransfo} & 82.1 & 80.7 & - & -   \\
\hline
\citep{chen_noetic} & - & - & 57.3 & 64.5 \\
\hline
\hline
 \multicolumn{5}{|l|}{\textbf{1 Attention Code}}  \\
\hline
Learnt-codes	& 81.9 $\pm$ 0.3 &	81.0 $\pm$ 0.1 & 56.2 $\pm$ 0.1 &	66.9 $\pm$ 0.7\\
First $m$ outputs	& 83.2 $\pm$ 0.2 &	81.5 $\pm$ 0.1 & 56.4 $\pm$ 0.3 &	66.8 $\pm$ 0.7 \\

Last $m$ outputs	& 82.9 $\pm$ 0.1 &  81.0 $\pm$ 0.1 & 56.1 $\pm$ 0.4 &	67.2 $\pm$ 1.1\\
Last $m$ outputs and $h^1_{ctxt}$ & - & - & - & -\\		

\hline
 \multicolumn{5}{|l|}{\textbf{4 Attention Codes}}  \\
\hline
Learnt-codes	& \textbf{83.8 $\pm$ 0.2} &	\textbf{82.2 $\pm$ 0.5} & 56.5 $\pm$ 0.5	& 66.8 $\pm$ 0.7 \\
First $m$ outputs	& 83.4 $\pm$ 0.2 &	81.6 $\pm$ 0.1 & \textbf{56.9 $\pm$ 0.5}	& \textbf{67.2 $\pm$ 1.3} \\
Last $m$ outputs	& 82.8 $\pm$ 0.2 &	81.3 $\pm$ 0.4 & 56.0 $\pm$ 0.5	& 65.8 $\pm$ 0.5\\
Last $m$ outputs and $h^1_{ctxt}$ &	82.9 $\pm$ 0.1	 & 81.4 $\pm$ 0.2 & 55.8 $\pm$ 0.3	& 66.1 $\pm$ 0.8\\

\hline
 \multicolumn{5}{|l|}{\textbf{16 Attention Codes}}  \\
\hline		
Learnt-codes &	84.4 $\pm$ 0.1 &	83.2 $\pm$ 0.1  & \textbf{57.7 $\pm$ 0.2} & \textbf{	67.8 $\pm$ 0.3}\\
First $m$ outputs &	\textbf{85.2 $\pm$ 0.1}	& \textbf{83.9 $\pm$ 0.2}  & 56.1 $\pm$ 1.7 & 	66.8 $\pm$ 1.1\\
Last $m$ outputs &	83.9 $\pm$ 0.2	 & 82.0 $\pm$ 0.4  & 56.1 $\pm$ 0.3 & 	66.2 $\pm$ 0.7\\
Last $m$ outputs and $h^1_{ctxt}$ &	83.8 $\pm$ 0.3 &	81.7 $\pm$ 0.3  & 56.1 $\pm$ 0.3 & 	66.6 $\pm$ 0.2\\

\hline
 \multicolumn{5}{|l|}{\textbf{64 Attention Codes}}  \\
\hline		
Learnt-codes &	84.9 $\pm$ 0.1 &	83.7 $\pm$ 0.2  & \textbf{58.3 $\pm$ 0.4} & 	67.0 $\pm$ 0.9\\
First $m$ outputs &	\textbf{86.0 $\pm$ 0.2} &	\textbf{84.2 $\pm$ 0.2} & 57.7 $\pm$ 0.6 & 	\textbf{67.1 $\pm$ 0.1} \\
Last $m$ outputs &	84.9 $\pm$ 0.3 &	82.9 $\pm$ 0.2  & 57.0 $\pm$ 0.2 & 	66.5 $\pm$ 0.5\\
Last $m$ outputs and $h^1_{ctxt}$ &	85.0 $\pm$ 0.2 &	83.2 $\pm$ 0.2  & 57.3 $\pm$ 0.3 & 	\textbf{67.1 $\pm$ 0.5}\\

\hline
 \multicolumn{5}{|l|}{\textbf{360 Attention Codes}}  \\
\hline		
Learnt-codes &	85.3 $\pm$ 0.3 &	83.7 $\pm$ 0.2 & 57.7 $\pm$ 0.3 & 	\textbf{68.9 $\pm$ 0.4} \\
First $m$ outputs &	\textbf{86.3 $\pm$ 0.1} &	84.6 $\pm$ 0.3 & 58.1 $\pm$ 0.4 & 	66.8 $\pm$ 0.7 \\
Last $m$ outputs &	\textbf{86.3 $\pm$ 0.1} &	\textbf{84.7 $\pm$ 0.3}  & 58.0 $\pm$ 0.4 & 	68.1 $\pm$ 0.5\\
Last $m$ outputs and $h^1_{ctxt}$ &	86.2 $\pm$ 0.3 &	84.5 $\pm$ 0.4 & \textbf{58.3 $\pm$ 0.4 }& 	68.0 $\pm$ 0.8 \\

\hline
\end{tabular}
\caption{Validation and test performance of Poly-encoder variants, with weights initialized from \citep{devlin-etal-2019-bert}. Scores are shown for ConvAI2 and  DSTC 7 Track 1. Bold numbers indicate the highest performing variant within that number of codes.
\label{table:all_poly_types}
}

\end{table*}

\begin{table*}
\center
\begin{small}
\begin{tabular}{|c|c|c|c|c|}
\hline
 & \multicolumn{4}{c|}{Scoring time (ms)}\\
\hline
 & \multicolumn{2}{c|}{CPU} & \multicolumn{2}{c|}{GPU}\\
\hline
Candidates & 1k  & 100k& 1k  & 100k \\
\hline
\hline
Bi-encoder & 115 & 160 & 19 & 22\\
\hline
Poly-encoder (First $m$ outputs) 16 & 119 & 551 & 17 & 37 \\
\hline
Poly-encoder (First $m$ outputs) 64 & 124 & 570 & 17 & 39 \\
\hline
Poly-encoder (First $m$ outputs) 360 & 120 & 619 & 17 &  45\\
\hline
Poly-encoder (Learnt-codes) 16 & 122 & 678 & 18 & 38 \\
\hline
Poly-encoder (Learnt-codes) 64 & 126 & 692 & 23 & 46 \\
\hline
Poly-encoder (Learnt-codes) 360 & 160 & 837 & 57 & 88 \\
\hline
Cross-encoder & 21.7k & 2.2M* & 2.6k & 266k*\\

\hline

\end{tabular}
\end{small}
\caption{Average time in milliseconds to predict the next dialogue utterance from $N$ possible candidates. * are inferred.
\label{all_timing}
}

\end{table*}

\newpage 

\begin{table*}[h]
\center
\setlength\tabcolsep{3.2pt} 
\scriptsize
\resizebox{\textwidth}{!}{\begin{tabular}{|l|c|c|c|c|c|c|c|c|c|c|}
\hline
 Dataset & \multicolumn{2}{c|}{ConvAI2} & \multicolumn{4}{c|}{DSTC 7} & \multicolumn{4}{c|}{Ubuntu v2} \\
 \hline
 split & dev & test & dev & \multicolumn{3}{c|}{test} & dev & \multicolumn{3}{c|}{test} \\
 \hline
metric & R@1/20 & R@1/20 & R@1/100 & R@1/100 & R@10/100 & MRR & R@1/10 & R@1/10 & R@5/10 & MRR\\
\hline
\hline
Hugging Face  & \multirow{ 2}{*}{82.1}  & \multirow{ 2}{*}{80.7} & \multirow{ 2}{*}{-} & \multirow{ 2}{*}{-} & \multirow{ 2}{*}{-} & \multirow{ 2}{*}{-} & \multirow{ 2}{*}{-} & \multirow{ 2}{*}{-} & \multirow{ 2}{*}{-} & \multirow{ 2}{*}{-} \\
\citep{wolf2019transfertransfo} & & & & & & & & & &  \\
\hline
\citep{chen_noetic} & - & - & 57.3 & 64.5 &  90.2 & 73.5 & -  & -  & -  & - \\
\hline
\citep{dong_et_hal_ubuntu2} & - & - & -  & -  & -  & -  & -  &  75.9 &	97.3 & 84.8 \\
\hline
\hline
 \multicolumn{11}{|l|}{pre-trained weights from \citep{devlin-etal-2019-bert} - Toronto Books + Wikipedia}  \\
\hline
Bi-encoder & 83.3 $\pm$ 0.2 & 81.7 $\pm$ 0.2 & 56.5 $\pm$ 0.4 & 66.8 $\pm$ 0.7 & 89.0 $\pm$ 1.0 & 74.6 $\pm$ 0.5 & 80.9 $\pm$ 0.6 & 80.6 $\pm$ 0.4 & 98.2 $\pm$ 0.1 & 88.0 $\pm$ 0.3\\ \hline 





Poly-encoder (First-m) 16 & 85.2 $\pm$ 0.1 & 83.9 $\pm$ 0.2 & 56.7 $\pm$ 0.2 & 67.0 $\pm$ 0.9 & 88.8 $\pm$ 0.3 & 74.6 $\pm$ 0.6 & 81.7 $\pm$ 0.5 & 81.4 $\pm$ 0.6 & 98.2 $\pm$ 0.1 & 88.5 $\pm$ 0.4\\ \hline 

Poly-encoder (Learnt-m) 16 & 84.4 $\pm$ 0.1 & 83.2 $\pm$ 0.1 & 57.7 $\pm$ 0.2 & 67.8 $\pm$ 0.3 & 88.6 $\pm$ 0.2 & 75.1 $\pm$ 0.2 & 81.5 $\pm$ 0.1 & 81.2 $\pm$ 0.2 & 98.2 $\pm$ 0.0 & 88.3 $\pm$ 0.1\\ \hline 

Poly-encoder (First-m) 64 & 86.0 $\pm$ 0.2 & 84.2 $\pm$ 0.2 & 57.1 $\pm$ 0.2 & 66.9 $\pm$ 0.7 & 89.1 $\pm$ 0.2 & 74.7 $\pm$ 0.4 & 82.2 $\pm$ 0.6 & 81.9 $\pm$ 0.5 & 98.4 $\pm$ 0.0 & 88.8 $\pm$ 0.3\\ \hline 

Poly-encoder (Learnt-m) 64 & 84.9 $\pm$ 0.1 & 83.7 $\pm$ 0.2 & 58.3 $\pm$ 0.4 & 67.0 $\pm$ 0.9 & 89.2 $\pm$ 0.2 & 74.7 $\pm$ 0.6 & 81.8 $\pm$ 0.1 & 81.3 $\pm$ 0.2 & 98.2 $\pm$ 0.1 & 88.4 $\pm$ 0.1\\ \hline 

Poly-encoder (First-m) 360 & 86.3 $\pm$ 0.1 & 84.6 $\pm$ 0.3 & 57.8 $\pm$ 0.5 & 67.0 $\pm$ 0.5 & 89.6 $\pm$ 0.9 & 75.0 $\pm$ 0.6 & 82.7 $\pm$ 0.4 & 82.2 $\pm$ 0.6 & 98.4 $\pm$ 0.1 & 89.0 $\pm$ 0.4\\ \hline 

Poly-encoder (Learnt-m) 360 & 85.3 $\pm$ 0.3 & 83.7 $\pm$ 0.2 & 57.7 $\pm$ 0.3 & 68.9 $\pm$ 0.4 & 89.9 $\pm$ 0.5 & 76.2 $\pm$ 0.2 & 81.5 $\pm$ 0.1 & 80.9 $\pm$ 0.1 & 98.1 $\pm$ 0.0 & 88.1 $\pm$ 0.1\\ \hline 

Cross-encoder & 87.1 $\pm$ 0.1& 84.8 $\pm$ 0.3 & 59.4 $\pm$ 0.4&   67.4 $\pm$ 0.7& 90.5 $\pm$ 0.3&  75.6 $\pm$ 0.4 &  83.3 $\pm$ 0.4& 82.8 $\pm$ 0.3& 98.4 $\pm$ 0.1& 89.4 $\pm$ 0.2\\
\hline
\hline
 \multicolumn{11}{|l|}{Our pre-training on Toronto Books + Wikipedia}  \\
 \hline
Bi-encoder & 84.6 $\pm$ 0.1 & 82.0 $\pm$ 0.1 & 54.9 $\pm$ 0.5 & 64.5 $\pm$ 0.5 & 88.1 $\pm$ 0.2 & 72.6 $\pm$ 0.4 & 80.9 $\pm$ 0.5 & 80.8 $\pm$ 0.5 & 98.4 $\pm$ 0.1 & 88.2 $\pm$ 0.4\\ \hline 

Poly-encoder (First-m) 16 & 84.1 $\pm$ 0.2 & 81.4 $\pm$ 0.2 & 53.9 $\pm$ 2.7 & 63.3 $\pm$ 2.9 & 87.2 $\pm$ 1.5 & 71.6 $\pm$ 2.4 & 80.8 $\pm$ 0.5 & 80.6 $\pm$ 0.4 & 98.4 $\pm$ 0.1 & 88.1 $\pm$ 0.3\\ \hline 

Poly-encoder (Learnt-m) 16 & 85.4 $\pm$ 0.2 & 82.7 $\pm$ 0.1 & 56.0 $\pm$ 0.4 & 65.3 $\pm$ 0.9 & 88.2 $\pm$ 0.7 & 73.2 $\pm$ 0.7 & 84.0 $\pm$ 0.1 & 83.4 $\pm$ 0.2 & 98.7 $\pm$ 0.0 & 89.9 $\pm$ 0.1\\ \hline 

Poly-encoder (First-m) 64 & 86.1 $\pm$ 0.4 & 83.9 $\pm$ 0.3 & 55.6 $\pm$ 0.9 & 64.3 $\pm$ 1.5 & 87.8 $\pm$ 0.4 & 72.5 $\pm$ 1.0 & 80.9 $\pm$ 0.6 & 80.7 $\pm$ 0.6 & 98.4 $\pm$ 0.0 & 88.2 $\pm$ 0.4\\ \hline 

Poly-encoder (Learnt-m) 64 & 85.6 $\pm$ 0.1 & 83.3 $\pm$ 0.1 & 56.2 $\pm$ 0.4 & 65.8 $\pm$ 0.7 & 88.4 $\pm$ 0.3 & 73.5 $\pm$ 0.5 & 84.0 $\pm$ 0.1 & 83.4 $\pm$ 0.1 & 98.7 $\pm$ 0.0 & 89.9 $\pm$ 0.0\\ \hline 

Poly-encoder (First-m) 360 & 86.6 $\pm$ 0.3 & 84.4 $\pm$ 0.2 & 57.5 $\pm$ 0.4 & 66.5 $\pm$ 1.2 & 89.0 $\pm$ 0.5 & 74.4 $\pm$ 0.7 & 81.3 $\pm$ 0.6 & 81.1 $\pm$ 0.4 & 98.4 $\pm$ 0.2 & 88.4 $\pm$ 0.3\\ \hline 

Poly-encoder (Learnt-m) 360 & 86.1 $\pm$ 0.1 & 83.8 $\pm$ 0.1 & 56.5 $\pm$ 0.8 & 65.8 $\pm$ 0.7 & 88.5 $\pm$ 0.6 & 73.6 $\pm$ 0.6 & 84.2 $\pm$ 0.2 & 83.7 $\pm$ 0.0 & 98.7 $\pm$ 0.1 & 90.1 $\pm$ 0.0\\ \hline 

Cross-encoder & 87.3 $\pm$ 0.5 & 84.9 $\pm$ 0.3 & 57.7 $\pm$ 0.5 & 65.3 $\pm$ 1.0 & 89.7 $\pm$ 0.5 & 73.8 $\pm$ 0.6 & 83.2 $\pm$ 0.8 & 83.1 $\pm$ 0.7 & 98.7 $\pm$ 0.1 & 89.7 $\pm$ 0.5\\ \hline 

\hline
 \multicolumn{11}{|l|}{Our pre-training on Reddit}  \\
 \hline
Bi-encoder & 86.9 $\pm$ 0.1 & 84.8 $\pm$ 0.1 & 60.1 $\pm$ 0.4 & 70.9 $\pm$ 0.5 & 90.6 $\pm$ 0.3 & 78.1 $\pm$ 0.3 & 83.7 $\pm$ 0.7 & 83.6 $\pm$ 0.7 & 98.8 $\pm$ 0.1 & 90.1 $\pm$ 0.4\\ \hline 

Poly-encoder (First-m) 16 & 89.0 $\pm$ 0.1 & 86.4 $\pm$ 0.3 & 60.4 $\pm$ 0.3 & 70.7 $\pm$ 0.7 & 91.0 $\pm$ 0.4 & 78.0 $\pm$ 0.5 & 84.3 $\pm$ 0.3 & 84.3 $\pm$ 0.2 & 98.9 $\pm$ 0.0 & 90.5 $\pm$ 0.1\\ \hline 

Poly-encoder (Learnt-m) 16 & 88.6 $\pm$ 0.3 & 86.3 $\pm$ 0.3 & 61.1 $\pm$ 0.4 & 71.6 $\pm$ 0.6 & 91.3 $\pm$ 0.3 & 78.4 $\pm$ 0.4 & 86.1 $\pm$ 0.1 & 86.0 $\pm$ 0.1 & 99.0 $\pm$ 0.1 & 91.5 $\pm$ 0.1\\ \hline 

Poly-encoder (First-m) 64 & 89.5 $\pm$ 0.1 & 87.3 $\pm$ 0.2 & 61.0 $\pm$ 0.4 & 70.9 $\pm$ 0.6 & 91.5 $\pm$ 0.5 & 78.0 $\pm$ 0.3 & 84.0 $\pm$ 0.4 & 83.9 $\pm$ 0.4 & 98.8 $\pm$ 0.0 & 90.3 $\pm$ 0.3\\ \hline 

Poly-encoder (Learnt-m) 64 & 89.0 $\pm$ 0.1 & 86.5 $\pm$ 0.2 & 60.9 $\pm$ 0.6 & 71.2 $\pm$ 0.8 & 91.3 $\pm$ 0.4 & 78.2 $\pm$ 0.7 & 86.2 $\pm$ 0.1 & 85.9 $\pm$ 0.1 & 99.1 $\pm$ 0.0 & 91.5 $\pm$ 0.1\\ \hline 

Poly-encoder (First-m) 360 & 90.0 $\pm$ 0.1 & 87.3 $\pm$ 0.1 & 61.1 $\pm$ 1.9 & 70.9 $\pm$ 2.1 & 91.5 $\pm$ 0.9 & 77.9 $\pm$ 1.6 & 84.8 $\pm$ 0.5 & 84.6 $\pm$ 0.5 & 98.9 $\pm$ 0.1 & 90.7 $\pm$ 0.3\\ \hline 

Poly-encoder (Learnt-m) 360 & 89.2 $\pm$ 0.1 & 86.8 $\pm$ 0.1 & 61.2 $\pm$ 0.2 & 71.4 $\pm$ 1.0 & 91.1 $\pm$ 0.3 & 78.3 $\pm$ 0.7 & 86.3 $\pm$ 0.1 & 85.9 $\pm$ 0.1 & 99.1 $\pm$ 0.0 & 91.5 $\pm$ 0.0\\ \hline 

Cross-encoder & 90.3 $\pm$ 0.2& 87.9 $\pm$ 0.2 & 63.9 $\pm$ 0.3&  71.7 $\pm$ 0.3&  92.4 $\pm$ 0.5& 79.0 $\pm$ 0.2 & 86.7 $\pm$ 0.1&  86.5 $\pm$ 0.1&99.1 $\pm$ 0.0&  91.9 $\pm$ 0.0\\ \hline 

\end{tabular}}
\caption{Validation and test performances of Bi-, Poly- and Cross-encoders. Scores are shown for ConvAI2, DSTC7 Track 1 and Ubuntu v2, and the previous state-of-the-art models in the literature.}
\label{table:best_scores_full}
\end{table*}

\end{document}